\documentclass{ecai} 

\usepackage{latexsym}
\usepackage{amssymb}
\usepackage{amsmath}
\usepackage{amsthm}
\usepackage{booktabs}
\usepackage{enumitem}
\usepackage{graphicx}
\usepackage{color}

\usepackage{multirow}
\usepackage{float}
\usepackage[normalem]{ulem}
\useunder{\uline}{\ul}{}
\usepackage{amsmath}
\usepackage{tikz}
\usetikzlibrary{automata, arrows, positioning, calc}
\pgfdeclarelayer{bg}    
\pgfsetlayers{bg,main}  

\newcommand{\vx}{\mathbf{x}}
\newcommand{\vh}{\mathbf{h}}

\DeclareMathOperator*{\argmin}{arg\,min}

\newcommand{\BibTeX}{B\kern-.05em{\sc i\kern-.025em b}\kern-.08em\TeX}

\begin{document}


\begin{frontmatter}


\paperid{1444} 


\title{TabResFlow: A Normalizing Spline Flow Model for \\ Probabilistic Univariate Tabular Regression}

\author[A,C]{\fnms{Kiran}~\snm{Madhusudhanan}\thanks{Corresponding Author. Email: kiranmadhusud@ismll.de}}
\author[A,C]{\fnms{Vijaya Krishna}~\snm{Yalavarthi}}
\author[B,C]{\fnms{Jonas}~\snm{Sonntag}} 
\author[A,C]{\fnms{Maximilian}~\snm{Stubbemann}}
\author[A,C]{\fnms{Lars}~\snm{Schmidt-Thieme}}

\address[A]{Information Systems and Machine Learning Lab (ISMLL), University of Hildesheim}
\address[B]{Volkswagen Financial Services AG}
\address[C]{VWFS Data Analytics Research Center}


\begin{abstract}
Tabular regression is a well-studied problem with numerous industrial applications, yet 
most existing approaches focus on point estimation, often leading to overconfident predictions. 
This issue is particularly critical in industrial automation, where trustworthy decision-making is essential. 
Probabilistic regression models address this challenge by modeling prediction uncertainty. However, 
many conventional methods assume a fixed-shape distribution (typically Gaussian), and resort to estimating distribution parameters. 
This assumption is often restrictive, as real-world target distributions can be highly complex.

To overcome this limitation, we introduce TabResFlow, a Normalizing Spline Flow model designed specifically 
for univariate tabular regression, where commonly used simple flow networks like RealNVP and Masked Autoregressive Flow 
(MAF) are unsuitable. TabResFlow consists of three key components:
(1) An MLP encoder for each numerical feature.
(2) A fully connected ResNet backbone for expressive feature extraction.
(3) A conditional spline-based normalizing flow for flexible and tractable density estimation.

We evaluate TabResFlow on nine public benchmark datasets, demonstrating that it consistently surpasses existing probabilistic regression models on likelihood scores. 
Our results demonstrate 9.64\% improvement compared to the strongest probabilistic regression model (TreeFlow), 
and on average 5.6 times speed-up in inference time compared to the strongest deep 
learning alternative (NodeFlow). Additionally, we validate the practical applicability of TabResFlow in a real-world used car price 
prediction task under selective regression. To measure performance in this setting, we introduce a novel Area Under Risk Coverage (AURC) metric 
and show that TabResFlow achieves superior results across this metric.
\end{abstract}

\end{frontmatter}


\section{Introduction}

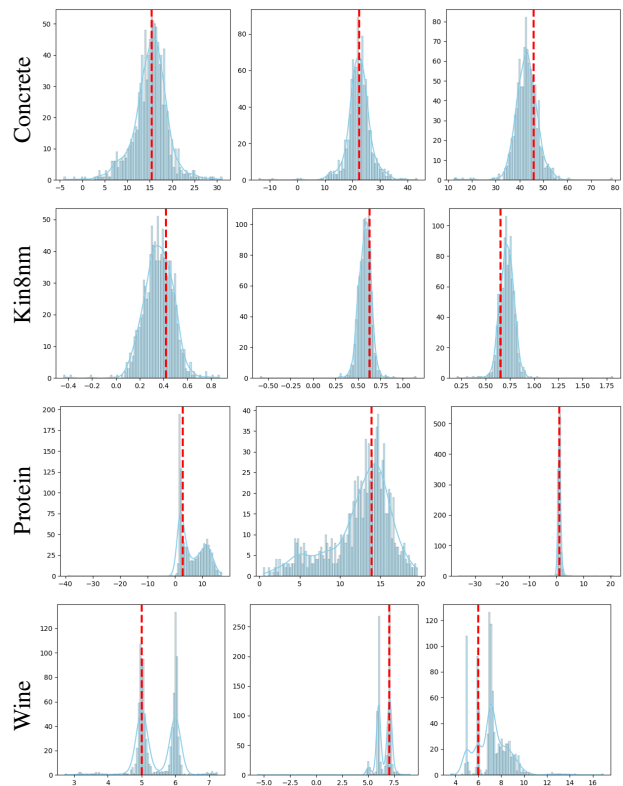
\begin{figure}[t]
	\begin{tikzpicture}[remember picture, overlay]
		\node[rotate=90, anchor=west] at (0.25, -2.0) {{{Concrete}}};
		\node[rotate=90, anchor=west] at (0.25, -4.4) {{{Kin8nm}}};
		\node[rotate=90, anchor=west] at (0.25, -7.2) {{{Protein}}};
		\node[rotate=90, anchor=west] at (0.25, -9.8) {{{Wine}}};
	\end{tikzpicture}
	
	\centering
	\includegraphics[width=0.90\columnwidth]{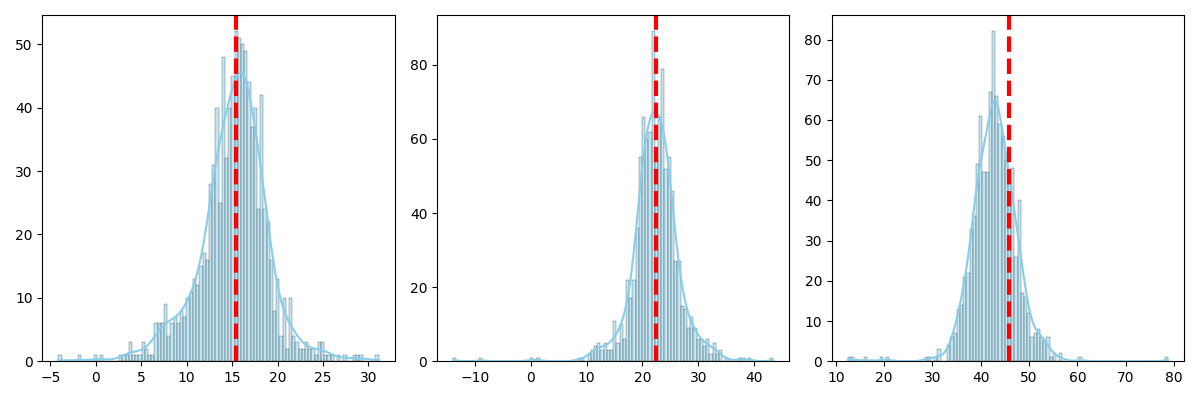}
	\label{fig:figure2}
	\includegraphics[width=0.90\columnwidth]{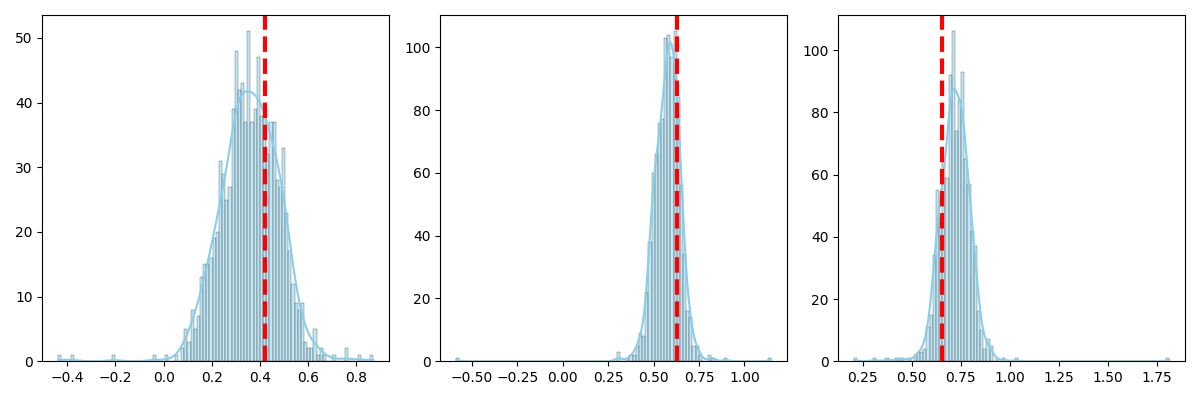}
	\label{fig:figure3}
	\includegraphics[width=0.90\columnwidth]{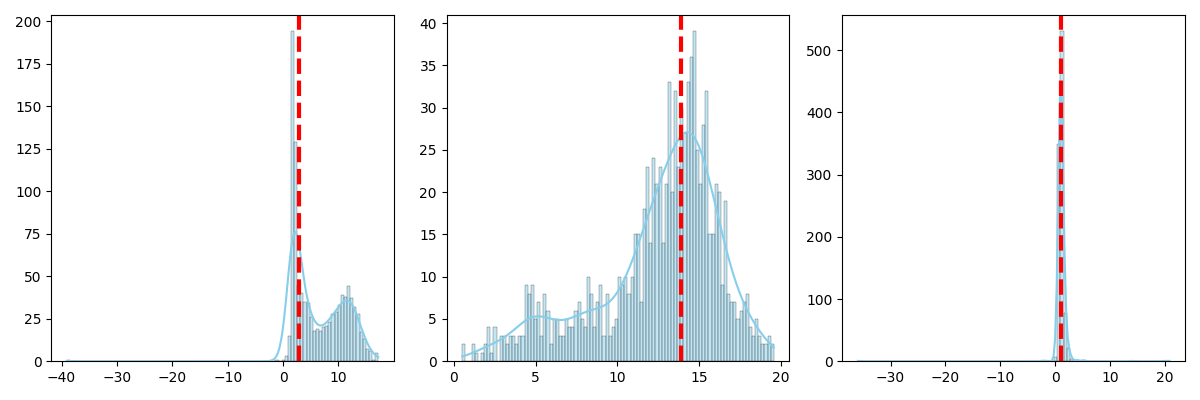}
	\label{fig:figure4}
	\includegraphics[width=0.90\columnwidth]{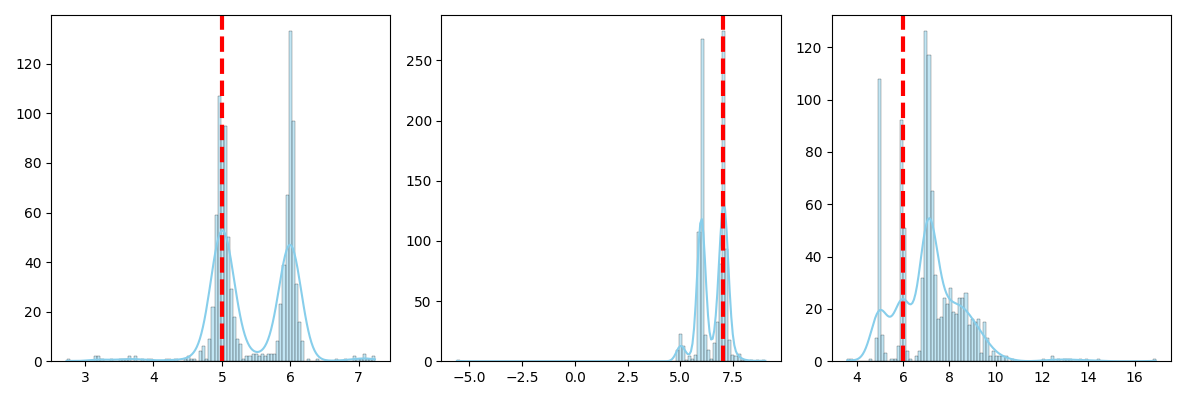}
	\label{fig:figure5}
	\caption{Comparison of the learned distribution from TabResFlow (in blue) with the ground truth (in red) for Concrete, Kin8nm, Protein and Wine Datasets. Protein and Wine follows a non-gaussian, multi-modal distribution.}
	\label{fig:data_distribution}
\end{figure}

Learning from tabular data remains one of the most extensively studied areas in machine learning, with tree-based boosting methods, such as XGBoost~\cite{chen2016xgboost}, CatBoost~\cite{prokhorenkova2018catboost}, and LightGBM~\cite{ke2017lightgbm}, serving as the de facto standard. Recently, however, deep learning approaches have begun to outperform boosting models and shown promising results that suggest a potential paradigm shift in tabular data modeling~\cite{somepalli2021saint,hollmann2023tabpfn}. Most of this progress has centered on classification tasks or point regression \cite{tabsurvey}, where the model predicts a single value, typically the expected value (for MSE) or the median (for MAE). In contrast, \textbf{probabilistic regression} aims to predict a full probability distribution over continuous target variables. This richer output enables more fine grained decision-making, especially in domains where understanding uncertainty is crucial.

Many practical applications require univariate tabular regression, where a single continuous outcome is predicted from structured input features. For instance, consider the task of predicting used-car prices: while an expected price estimate is informative, access to the entire predictive distribution provides deeper insight into uncertainty. Low-uncertainty predictions can be automated confidently, while high-uncertainty ones may be deferred for human review, a strategy commonly used in selective prediction. Similar needs arise in domains such as healthcare \cite{lipkova2019personalized}, customer behavior modeling \cite{dew2024probabilistic}, and industrial quality control \cite{mehdiyev2024quantifying}.

Despite its utility, univariate probabilistic regression on tabular data remains relatively underexplored. Existing methods often make strong distributional assumptions, such as Gaussian on the target variable~\cite{madhusudhanan2024probsaint,duan2020ngboost,HypertuningMLPs}, which limits their expressiveness. For instance, 
Figure \ref{fig:data_distribution} illustrates the learned distribution across four real-world tabular datasets. 
Assuming a Gaussian distribution to model such datasets would be inaccurate.
Normalizing flows (NFs)~\cite{papamakarios2021normalizing}, such as RealNVP~\cite{dinh2017density} and Masked Autoregressive Flows (MAFs)~\cite{papamakarios2017masked}, offer a more flexible alternative for modeling complex, non-Gaussian distributions. However, these techniques have primarily been applied in multivariate settings and, to the best of our knowledge, cannot be directly implemented for univariate probabilistic regression tasks due to their reliance on model constructions that ensure flexible invertibility and tractable Jacobian computation.

Recent models like TreeFlow~\cite{wielopolski2023treeflow} and NodeFlow~\cite{wielopolski2024nodeflow} attempt to bridge this gap by combining tree-based or NODE-style \cite{Popov2019NeuralOD} encoders with Continuous Normalizing Flows (CNFs) \cite{yang2019pointflow}. While expressive, these models suffer from notable drawbacks. TreeFlow scales poorly due to its dependence on encoding the entire CatBoost model as a conditioning network, and both TreeFlow and NodeFlow rely on CNFs, which involve solving differential equations during training and inference. This results in significant computational overhead, limiting their practicality for large-scale deployment. While flow-matching \cite{Lipman2022FlowMF} attempts to address the training limitations, they are still expensive at the time of inference.

To address these challenges, we propose \emph{TabResFlow}, a ResNet-based architecture tailored for univariate probabilistic tabular regression. TabResFlow simplifies NodeFlow by replacing its complex NODE-based backbone with a residual multi-layer perceptron (MLP), and it replaces the expensive CNF component with a more efficient Rational Quadratic Neural Spline Flow (RQ-NSF)~\cite{durkan2019neural}. This design choice is more flexible and also fast during inference.

Negative Log-likelihood is the standard metric to evaluate probabilistic predictions. For use-cases such as selective regression, we propose \emph{Area Under the Risk-Coverage Curve} (AURC)~\cite{geifman2017selective} as a complementary evaluation metric. AURC assesses model uncertainty by ranking predictions according to confidence and measuring error across different coverage levels. We validate our approach through extensive experiments on benchmark datasets and on a large-scale, real-world used-car dataset. TabResFlow consistently outperforms existing methods, achieving both higher accuracy and inference time speed ups.
Our key contributions are as follows:

%
%
\begin{itemize}
\item We introduce \textbf{TabResFlow}, a ResNet-style MLP architecture combined with a monotonic rational quadratic spline flow, enabling the modeling of expressive univariate distributions for probabilistic regression. 
\item For selective regression tasks requiring reliable predictive uncertainty, we propose \emph{Area Under the Risk-Coverage Curve (AURC)} as a practically interpretable evaluation metric. 
\item TabResFlow achieves \textbf{state-of-the-art performance} on benchmark datasets and on a real-world, large-scale, used-car pricing case study. We provide our source code: \texttt{https://github.com/18kiran12/TabResFlow}.
\end{itemize}

\section{Related Works}

\paragraph{\textbf{Tabular Regression.}} Traditionally, Gradient Boosting Decision Tree (GBDT) models have been widely recognized as 
the top choice for tabular regression, capable of generating accurate predictions at a fraction of the 
runtime required by deep learning models. Especially,  XGBoost \cite{chen2016xgboost}, LightGBM \cite{ke2017lightgbm}, 
and CatBoost \cite{prokhorenkova2018catboost} are well-known 
for their high accuracy, reliability and efficiency. Recent advances in Deep Learning for tabular data 
have benefited with the use of column and row attention like SAINT \cite{somepalli2021saint} and
TabPFN \cite{hollmann2023tabpfn}. However, the expensive dual stage attention in SAINT and TabPFN 
restricts these methods to smaller datasets.   
Another line of work investigates Neural Oblivious Decision Ensembles
or NODE \cite{Popov2019NeuralOD} as an end-to-end gradient-based trainable alternate to GBDT models. 
However, recent literature comparing the NODE with a simple ResNet architecture demonstrates that a 
simple fully-connected ResNet architecture achieves a better performance than the NODE architecture \cite{gorishniy2021revisiting}. 
Furthermore, in \cite{gorishniy2022embeddings}, the authors investigate the use of numerical embeddings 
for tabular data and show that a ResNet architecture with numerical embeddings achieve 
similar performance to recent Transformer architectures for tabular data. 
Aforementioned studies recommends the use of a simple ResNet architecture with numerical embeddings 
as an efficient feature extractor for tabular regression.

\paragraph{\textbf{Probabilistic Tabular Regression.}} Although XGBoost, CatBoost and LightGBM 
excel in point prediction accuracy, incorporating probabilistic outputs often require post-hoc
calibrations or modifications as in \cite{marz2019xgboostlss}. A more principled GBDT based 
probabilistic prediction is offered by the NGBoost method \cite{duan2020ngboost}, that directly estimates
the probabilistic distribution by incorporating a probabilistic loss function and learning 
using Natural Gradients in Riemannian space. A similar line of 
research is the PGBM \cite{sprangers2021probabilistic} algorithm, that learns the distribution 
parameters as the tree leaf weights, and generates probabilistic estimates from a single decision 
tree model. However, the above mentioned models 
either learn for a particular distribution or is tuned for a particular distribution after training.
Another interesting work relies on virtual ensembles from a CatBoost model to generate 
uncertainty from the ensembles \cite{malinin2020uncertainty}. However, these ensembles are never trained 
to predict the uncertainty and therefore could be prone to inaccurate uncertainties.

Flexible probabilistic modeling, which removes the need to know the underlying distribution in advance, is a key research area. Bayesian Neural Networks (BNNs)~\cite{hernandez2015probabilistic}, model weights as distributions for flexible distribution learning, but rely on choosing suitable priors and use approximate inference methods, making them computationally expensive. Mixture of Gaussians models~\cite{distFreeShi} offer another distribution-free approach by combining multiple Gaussian components, but they typically require many components to capture complex distributions effectively.

Normalizing Flows on the other hand is a popular choice to directly learn the 
true probability density of the target variable with the use of invertible differentiable transformations.
Nevertheless utilizing Normalizing Flows for the univariate setting is restricted to employing either Spline Flows \cite{durkan2019neural} or with the use of Continuous Normalizing Flows (CNF) \cite{yang2019pointflow}.
Other well known methods like RealNVP~\cite{dinh2017density} and Masked Autoregressive Flows (MAFs)~\cite{papamakarios2017masked} are incapable of modeling univariate targets.  
CNF's have previously been employed in TreeFlow \cite{wielopolski2023treeflow} and NodeFlow
\cite{wielopolski2024nodeflow} models, and these models can learn the true density 
without any underlying assumption on the target distribution. However, TreeFlow model is not
end-to-end trainable and not scalable to larger datasets, as the model requires the entire 
feature extractor Catboost model to be passed in as the conditioning for the CNF. Although, 
NodeFlow rectifies these shortcomings from the TreeFlow model by using the NODE architecture as the 
backbone feature extractor making the model end-to-end trainable, the CNF component still requires 
solving ordinary differential equations. This leads to 
significant increase in training and testing time complexity. Our work attempts to improve the 
prediction accuracy while reducing computational complexity, with the use of simple ResNet backbone and 
Neural Spline Flow based on Monotonic Rational Quadratic Splines as the normalizing flow 
transformations. 

\section{Problem Formulation} \label{sec:problem}

We consider the task of univariate probabilistic tabular regression, where the objective is to model the conditional distribution of a scalar target given a fixed-dimensional input vector.

Formally, let the training dataset consist of $N$ many samples of $(\vx, y) \sim p^\text{data}$, where each predictor $\vx \in \mathcal{X}^F$ and corresponding output $y \in \mathbb{R}$ are drawn i.i.d. from an unknown joint distribution $p^\text{data}(\vx, y)$.
We place no restrictions on the type of input features; that is, $\vx$ may include categorical and numerical attributes.
The goal is to learn a predictive model $\hat{p}(y \mid \vx)$ that estimates the true conditional distribution of $y$ given the input $\vx$.

The quality of the probabilistic prediction is evaluated via negative log-likelihood (NLL): $-\log \hat{p}(y \mid \vx)$.
The objective is to find a model $\hat{p}$ that minimizes this expected NLL, thereby aligning the predictive distribution as closely as possible with the true conditional distribution:
\begin{align} 
	\hat{p}^* = \argmin_{\hat{p}} \mathop\mathbb{E}_{(\vx, y) \sim p^\text{data}} \left[ -\log \hat{p}(y \mid \vx) \right]. 
\end{align}

\section{TabResFlow}

Often, models assume that the predictions follow a fixed shape distribution such as Gaussian distribution i.e. $y\sim \mathcal{N}(\mu(\vx), \sigma^2(\vx))$.
However, in real-world applications, there are many cases where the true conditional distribution $p(y\mid \vx)$ may not follow the Gaussian distribution.
Hence, assuming a Gaussian distribution to the target may not capture the true underlying distribution.
For this, we propose to use Normalizing Flows which does not assume any fixed shape to the underlying distribution and can predict any random distribution.

\subsection{Normalizing Flows for Probabilistic Regression}%
\label{sec:flows_for_regression}
A normalizing flow models the true probability distribution \( p_Y \) of a target variable \( y \in \mathbb{R}^K \) by transforming a simple base distribution \( p_Z \), typically a Gaussian, through a series of invertible transformations \( f: \mathbb{R}^K \to \mathbb{R}^K \). For univariate setting $K=1$.  
The density of \( y \) is given by the change-of-variables formula:
\begin{align}
	p_Y(y) = p_Z(f^{-1}(y)) \left| \det \left( \frac{\partial f^{-1}(y)}{\partial y} \right) \right|
\end{align}
A normalizing flow can be conditioned on predictors \( \vx \in \mathbb{R}^F \) by making the transformation \( f \) and the base distribution dependent on \( \vx \) as well:
\begin{align} \label{eq:condflow}
	p_Y(y\mid \vx) = p_Z(f^{-1}(y\mid \vx)\mid \vx) \left| \det \left( \frac{\partial f^{-1}(y\mid \vx)}{\partial y} \right) \right|
\end{align}

Widely used Flow architectures such as Real-NVP~\cite{dinh2017density} and Masked Auto Regressive Flows~\cite{papamakarios2017masked}, to the best of our knowledge, are applicable only to multivariate distributions due to the constraints on the transformation function.
On the other hand, Residual Flows~\cite{chen2019residual} and Continuous Normalizing Flows \cite{yang2019pointflow} can be used for the univariate distributions, however they suffer from numerical approximations of the inverse and the Jacobian of the transformation function $f$.
Hence, we propose to use spline functions~\cite{durkan2019neural}, which are univariate, have explicit inverse and Jacobian computation.

\paragraph{Univariate transformation via splines.}
Monotonic rational-quadratic splines (RQ-NSF)~\cite{durkan2019neural} provide a versatile and invertible univariate transformation for normalizing flows, allowing modeling of complex data distributions. These splines construct a piecewise rational-quadratic function across intervals defined by knots.

Formally, given a set of monotonically increasing points $\{(z^m, y^m)\}_{m=1:M}$ called knots, that is $z^m < z^{m+1}$ and $y^m < y^{m+1}$, along with their corresponding derivatives $\{\Delta_m > 0\}_{m=1:M}$, then the Monotonic rational-quadratic spline transformation $f(z)$ within a bin $z\in [z^m, z^{m+1}]$ is:
\begin{align}
	f(z) = y^m + \frac{(y^{m+1} - y^m)[s_m\zeta^2 + \Delta_m \zeta(1-\zeta)]}{s_m + [\Delta_{m+1} + \Delta_m - 2s_m]\zeta(1-\zeta)}
\end{align}
where $s_m = \frac{y^{m+1} - y^m}{z^{m+1} - z^m}$ is the slope between the knots and $\zeta(z) = \frac{z - z^m}{z^{m+1} - z^m}$ is the standard interpolation. The derivative of the function w.r.t. $z$ can be given as:
\begin{align*}
	\frac{\text{d}}{\text{d}z}f(z) = 
	\frac{(s_m^2)[\Delta_{m+1}\zeta^2 + 2s_m\zeta(1-\zeta) + \Delta_m(1-\zeta)^2]}{[s_m + [\Delta_{m+1} + \Delta_m - 2s_m]\zeta(1-\zeta)]^2}
\end{align*}
For the conditional spline transformation $f(z \mid \vh, \theta)$, the function parameters, such as the widths and heights of each bin and the derivatives at the knots, are computed from the conditioning input $\vh \in \mathbb{R}^{F \cdot D}$ using trainable parameters $\theta$.
We call $\vh$ is an encoding of the input features $\vx$ derived from a ResNet-based MLP architecture.

\subsection{Encoding Numerical and Categorical Features in Tabular Data}

In tabular data, categorical features are commonly embedded using a learnable lookup table, mapping each category index to a dense vector. In contrast, numerical features are often passed directly as raw scalars, potentially limiting the model's expressiveness.
Recent studies \cite{gorishniy2022embeddings,somepalli2021saint} show that applying shallow, non-linear embeddings to numerical inputs using MLPs can improve performance significantly, especially when combined with residual architectures.

In TabResFlow, each numerical feature $x_i$ is transformed via an MLP:
\begin{align*}
	\mathbf{e}_i = \text{Linear}(\text{ReLU}(\text{Linear}(x_i)))
\end{align*}
Categorical features are likewise embedded into $\mathbb{R}^D$ using a learned embedding matrix (e.g., nn.Embedding in PyTorch).

Given \( F \) features in total, each feature is transformed to a \( D \)-dimensional embedding. These embeddings form a matrix:
\[
\mathbf{E} = [\mathbf{e}_1, \ldots, \mathbf{e}_F]^\top \in \mathbb{R}^{F \times D}
\]
Flattening the embedding matrix yields:
\[
\mathbf{h}^{(0)} = \text{flatten}(\mathbf{E}) \in \mathbb{R}^{F \cdot D}
\]
which is used as input to the ResNet encoder.

\subsection{ResNet Encoder for Feature Extraction}

Transformer architectures \cite{somepalli2021saint,gorishniy2022embeddings} achieve state-of-the-art results but come with high computational costs. In contrast, ResNet, as demonstrated in \cite{gorishniy2022embeddings}, is a simple and efficient architecture that leverages residual connections to facilitate effective learning and attains comparable performance to Transformers. Therefore, we adopt the ResNet architecture \cite{gorishniy2022embeddings} for efficient representation learning from tabular data.
Each residual block in the network is defined as:
\begin{align}
	\text{ResNetBlock}(\mathbf{h}) = \mathbf{h} + \text{MLP}(\mathbf{h})
\end{align}
where:
\begin{align}
	\text{MLP}(\mathbf{h}) = \text{Linear}(\text{ReLU}(\text{Linear}(\text{BatchNorm}(\mathbf{h}))))
\end{align}
We stack \( L \) such blocks to obtain the final encoding:
\begin{align*}
	\vh = \mathbf{h}^{(L)} = \text{ResNetBlock}^{(L)} \circ \ldots \circ \text{ResNetBlock}^{(1)}(\mathbf{h}^{(0)})
\end{align*}
Dropout~\cite{srivastava14a} is applied after each Linear layer for regularization.

\begin{figure}[t]
	\centering
	\begin{tikzpicture}
	\node[minimum width=9mm] at (0,0) (xnum) {$x^\text{num}$};
	\node[above=0.5cm of xnum, minimum width=9mm] (xcat) {$x^\text{cat}$};
	\node[right=0.7cm of xnum, draw, minimum width=1.5cm, minimum height=0.7cm, rounded corners, fill=gray!40] (encmlp) {MLP};
	\node[right=0.7cm of xcat, draw, minimum width=1.5cm, minimum height=0.7cm, rounded corners, fill=gray!40] (enclin) {Linear};
	\coordinate(a) at ($(encmlp)!0.5!(enclin)$);
	\node[right=1.2cm of a, inner sep=0, label=60:$\mathbf{h}^{(0)}$] (+) {$\textcircled{\#}$};
	\node[right=0.7cm of +, draw, rounded corners, minimum height=0.7cm,label=10:{$\vh^{(L)}$}, fill=yellow!20] (resnet) {ResNet};
	\node[right=0.7cm of resnet, draw, rounded corners, minimum height=0.7cm, , fill=green!30] (flow) {Spline Flow};
	\node[above=0.5cm of flow] (y) {$y$};
	\node[below=0.5cm of flow, align=center] (z) {$z \sim p_Z$};
	
	\draw[-latex] (xcat) -- (enclin);
	\draw[-latex] (xnum) -- (encmlp);
	\draw[-latex] (enclin) -| (+);
	\draw[-latex] (encmlp) -| (+);
	\draw[-latex] (+) -- (resnet);
	\draw[-latex] (resnet) -- (flow);
	\draw[-latex] (y) -- (flow);
	\draw[-latex] (flow) -- (z);
\end{tikzpicture}
	
	\begin{tikzpicture}[node distance=3mm]
	\node at (0,0) (h0) {$\vh^{(0)}$};
	\node[draw, rounded corners,minimum height=0.7cm, above=of h0, fill=pink!40] (r1) {ResNetBlock};
	\node[above=of r1, inner sep=0] (p1) {$\oplus$};
	\node[above=0.5cm of p1, inner sep=0] (pL1) {$\oplus$};
	\node[draw, rounded corners,minimum height=0.7cm, above=of pL1, fill=pink!40] (rL) {ResNetBlock};
	\node[above=of rL, inner sep=0] (pL) {$\oplus$};
	\node[above=of pL] (hL) {$\vh^{(L)}$};
	\node[draw, rounded corners, minimum height=0.7cm,minimum width = 2cm, right=1.5cm of r1, fill=brown!20] (bnorm) {Batch Norm};
	\node[draw, rounded corners, minimum height=0.7cm,minimum width = 2cm, above=of bnorm, fill=gray!40] (lin1) {Linear};
	\node[draw, rounded corners, minimum height=0.7cm,minimum width = 2cm, above=of lin1, fill=brown!20] (relu) {relu};
	\node[draw, rounded corners, minimum height=0.7cm,minimum width = 2cm, above=of relu, fill=gray!40] (lin2) {Linear};

	\draw[-latex] (h0) -- (r1);
	\draw[-latex] (r1) -- (p1);
	\draw[-latex, dotted] (p1) -- (pL1);
	\draw[-latex] (pL1) -- (rL);
	\draw[-latex] (rL) -- (pL);
	\draw[-latex] (pL) -- (hL);
	\draw[-latex, rounded corners] (h0) -- +(4em, 0) |- (p1);
	\draw[-latex, rounded corners] ([yshift=1mm]pL1.north) -- +(4em, 0) |- (pL);
	\draw[-latex] (bnorm) -- (lin1);
	\draw[-latex] (lin1) -- (relu);
	\draw[-latex] (relu) -- (lin2);
	\draw[-latex] ([yshift=-3mm]bnorm.south) -- (bnorm);
	\draw[-latex] (lin2) -- ([yshift=3mm]lin2.north);
	\begin{pgfonlayer}{bg}    
		\coordinate(a) at ($(p1)!0.9!(pL1)$);
		\node[draw, rounded corners, fill=yellow!20, minimum height=5.2cm, minimum width=2.8cm, label=90:{ResNet}] at (a) (block) {};
		\coordinate(b) at ($(lin1)!0.5!(relu)$);
		\node[draw, rounded corners, fill=pink, fill opacity=0.4, minimum height=4.5cm, minimum width=2.8cm, label=90:ResNetBlock] at (b) (block2) {};
	\end{pgfonlayer}
	\node[below=3mm of h0] (dummy) {};
	
\end{tikzpicture}
	\caption{Architecture of TabResFlow for Probabilistic Tabular Regression. Dropout is used after each Linear 
		Layer in the ResNetBlock. $\#$ denotes the concatenation and flattening operation.}
	\vspace{2em}
	\label{fig:tabresflow}
\end{figure}
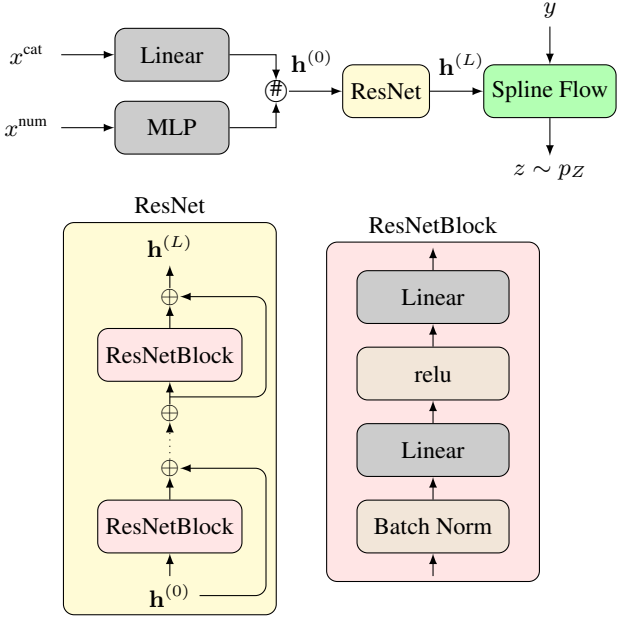

\subsection{Overall Architecture: TabResFlow}  
\label{sec:tabreflow}

The overall TabResFlow model, illustrated in Figure~\ref{fig:tabresflow}, comprises three main components: (1) encoding of numerical features, (2) a feature extraction module based on residual networks, and (3) conditional normalizing flows using RQ-NSF. These components collectively enable flexible and efficient modeling of the target distribution  \( p(y \mid \vx) \), leveraging the computational efficiency of the ResNet backbone and the straightforward quadratic inversions of RQ-NSF, while avoiding restrictive assumptions such as Gaussianity.

\paragraph{Training.} Given a batch $\mathcal{B}$ of instances, the model learns to transform the base distribution \( p_Z(z) \) into the empirical distribution of the target \( y \) by minimizing the conditional negative log-likelihood:
\[
\mathcal{L}(\theta) = \frac{1}{\vert \mathcal{B}\vert}\sum_{(\vx, y)\sim \mathcal{B}} -\log \hat{p}(y \mid \vh, \theta)
\]
$\hat{p}(y\mid \vh, \theta)$ is computed using Equation~\eqref{eq:condflow} where $\vh, \theta$ takes the role of $\vx$, and $\theta$ are the trainable parameters of spline function.
Since we use rational quadratic spline functions, the flow is invertible and have tractable Jacobians, the log-likelihood can be computed exactly and efficiently.

\paragraph{Inference.} TabResFlow provides a full predictive distribution \( \hat{p} \), not just a point estimate.
We can compute the exact density of $y: \hat{p}(y\mid \vx)$.
Further, we can also compute median or expected value which are used to evaluate the point estimation.
For example, to compute the expected value ($\mathbb{E}(y\mid \vx)$), we sample 1000 many samples and take the average value.
Median can be computed directly by transforming the median of base distribution to target distribution.


\section{Experiments}
\label{sec:Experiments}

We focus on univariate probabilistic regression, following the same experimental protocol outlined in \cite{wielopolski2024nodeflow}. Experiments are conducted on nine UCI regression datasets, as recommended in \cite{wielopolski2023treeflow,wielopolski2024nodeflow}, with dataset statistics summarized in Table \ref{tab:dataset_characteristics} in the supplementary material. We use 20 cross-validation splits for most datasets, except for Protein and YearMSD where we apply five and one split(s), respectively, due to their larger sizes.

Model performance is primarily assessed using negative log-likelihood (NLL) on the test data, which aligns with TabResFlow’s training objective. Root Mean Squared Error (RMSE) is also reported for point prediction accuracy following~\cite{wielopolski2023treeflow,wielopolski2024nodeflow}, while CRPS results for select baselines are included in the supplementary material \ref{app:crps_public}. Hyperparameters are optimized using Optuna, and training is performed on an NVIDIA 1080 GPU with 32 GB memory and a batch size of 2048.

\paragraph{\textbf{Baselines.}} Table \ref{tab:nll} presents a comparative 
evaluation of the proposed TabResFlow against relevant baseline methods on NLL. 
These include ensemble of simple Neural Networks (Deep. Ens.) \cite{lakshminarayanan2017simple}, CatBoost \cite{prokhorenkova2018catboost},
NGBoost \cite{duan2020ngboost}, PGBM \cite{sprangers2021probabilistic}, TreeFlow \cite{wielopolski2024nodeflow} and the more recent NodeFlow \cite{wielopolski2024nodeflow}.
While TabResFlow results are directly reported, baseline NLL scores are sourced from the findings of TreeFlow \cite{wielopolski2023treeflow} and NodeFlow \cite{wielopolski2024nodeflow}. This is possible as we use the exact same training and 
evaluation pipeline as in \cite{wielopolski2024nodeflow}.


\subsection{Results}

\begin{table*}[t]
	\caption{Comparison of probabilisitic tabular regression models on benchmark datasets for NLL score. The best value is represented in \textbf{bold}, whereas the 
		second best value is \underline{{underlined}}.}
	\label{tab:nll}
	\centering
		\begin{tabular}{lrrrrrrr}
			\midrule
			\textbf{Dataset} & \multicolumn{1}{c}{\textbf{Deep. Ens.}}       & \multicolumn{1}{c}{\textbf{CatBoost}} & \multicolumn{1}{c}{\textbf{NGBoost}}         & \multicolumn{1}{c}{\textbf{PGBM}}            & \multicolumn{1}{c}{\textbf{TreeFlow}}      & \multicolumn{1}{c}{\textbf{NodeFlow}}     & \multicolumn{1}{c}{\textbf{TabResFlow}}       \\ \midrule
			Concrete         & 3.06 \scriptsize{$\pm$ 0.18}           & 3.06 \scriptsize{$\pm$ 0.13}   & 3.04 \scriptsize{$\pm$ 0.17}         & \textbf{2.75 \scriptsize{$\pm$ 0.21}} & 3.02 \scriptsize{$\pm$ 0.15}        & 3.15 \scriptsize{$\pm$ 0.21}       & \underline{2.90 \scriptsize{$\pm$ 0.15}}     \\
			Energy           & 1.38 \scriptsize{$\pm$ 0.22}           & 1.24 \scriptsize{$\pm$ 1.28}   & \textbf{0.60 \scriptsize{$\pm$ 0.45}} & 1.74 \scriptsize{$\pm$ 0.04}          & 0.85 \scriptsize{$\pm$ 0.35}        & 0.90 \scriptsize{$\pm$ 0.25}       & \underline{0.77 \scriptsize{$\pm$ 0.19}}     \\
			Kin8nm           & \underline{-1.20 \scriptsize{$\pm$ 0.02}}    & -0.63 \scriptsize{$\pm$ 0.02}  & -0.49 \scriptsize{$\pm$ 0.02}         & -0.54 \scriptsize{$\pm$ 0.04}        & -1.03 \scriptsize{$\pm$ 0.06}       & -1.10 \scriptsize{$\pm$ 0.05}      & \textbf{-1.29 \scriptsize{$\pm$ 0.04}} \\
			Naval            & \textbf{-5.63 \scriptsize{$\pm$ 0.05}} & -5.39 \scriptsize{$\pm$ 0.04}  & -5.34 \scriptsize{$\pm$ 0.04}         & -3.44 \scriptsize{$\pm$ 0.04}         & \underline{-5.54 \scriptsize{$\pm$ 0.16}} & -5.45 \scriptsize{$\pm$ 0.08}      & -5.30 \scriptsize{$\pm$ 0.11}          \\
			Power            & 2.79 \scriptsize{$\pm$ 0.04}           & 2.72 \scriptsize{$\pm$ 0.12}   & 2.79 \scriptsize{$\pm$ 0.11}          & \textbf{2.60 \scriptsize{$\pm$ 0.02}} & 2.65 \scriptsize{$\pm$ 0.06}        & \underline{2.62 \scriptsize{$\pm$ 0.05}} & \textbf{2.60 \scriptsize{$\pm$ 0.04} } \\
			Protein          & 2.83 \scriptsize{$\pm$ 0.02}           & 2.73 \scriptsize{$\pm$ 0.07}   & 2.81 \scriptsize{$\pm$ 0.03}          & 2.79 \scriptsize{$\pm$ 0.01}          & \underline{2.02 \scriptsize{$\pm$ 0.02}}  & 2.04 \scriptsize{$\pm$ 0.04}       & \textbf{1.95 \scriptsize{$\pm$ 0.04}}  \\
			Wine             & 0.94 \scriptsize{$\pm$ 0.12}           & 0.93 \scriptsize{$\pm$ 0.08}   & 0.91 \scriptsize{$\pm$ 0.06}          & 0.97 \scriptsize{$\pm$ 0.20}          & \underline{-0.56 \scriptsize{$\pm$ 0.62}} & -0.21 \scriptsize{$\pm$ 0.28}      & \textbf{-0.85 \scriptsize{$\pm$ 0.27}} \\
			Yacht            & 1.18 \scriptsize{$\pm$ 0.21}           & 0.41 \scriptsize{$\pm$ 0.39}   & \underline{0.20 \scriptsize{$\pm$ 0.26}}    & \textbf{0.05 \scriptsize{$\pm$ 0.28}} & 0.72 \scriptsize{$\pm$ 0.40}        & 0.79 \scriptsize{$\pm$ 0.55}       & 0.67 \scriptsize{$\pm$ 0.32}           \\ 
			Year MSD         & 3.35 \scriptsize{$\pm$ NA }                            & 3.43 \scriptsize{$\pm$ NA }                          & 3.43 \scriptsize{$\pm$ NA }                         & 3.61 \scriptsize{$\pm$ NA }                                 & 3.27 \scriptsize{$\pm$ NA }                          & \textbf{3.09 \scriptsize{$\pm$ NA }}                 & \underline{3.21 \scriptsize{$\pm$ NA }} \\\midrule
		\end{tabular}%
\end{table*}

\begin{table*}[]
	\caption{Comparison of probabilisitic tabular regression models on benchmark datasets for RMSE score. The best value is represented in \textbf{bold}, whereas the 
		second best value is \underline{{underlined}}.}
	\label{tab:rmse}
	\centering
	\begin{tabular}{l|rrrrrrrr}
		\midrule
		\textbf{Dataset} & \multicolumn{1}{c}{\textbf{Deep. Ens.}}       & \multicolumn{1}{c}{\textbf{CatBoost}} & \multicolumn{1}{c}{\textbf{NGBoost}}         & \multicolumn{1}{c}{\textbf{PGBM}}            & \multicolumn{1}{c}{\textbf{TreeFlow}}      & \multicolumn{1}{c}{\textbf{NodeFlow}}     & \multicolumn{1}{c}{\textbf{TabResFlow}}       \\ \midrule
		Concrete & 6.03 \scriptsize{$\pm$ 0.58}       & 5.21 \scriptsize{$\pm$ 0.53}          & 5.06 \scriptsize{$\pm$ 0.61}        & \textbf{ 3.97 \scriptsize{$\pm$ 0.76}} & 5.41 \scriptsize{$\pm$ 0.72} & 5.51 \scriptsize{$\pm$ 0.66}       & {\ul 5.01 \scriptsize{$\pm$ 0.70}}     \\
		Energy   & 2.09 \scriptsize{$\pm$ 0.29}       & 0.57 \scriptsize{$\pm$ 0.06}          & {\ul 0.46 \scriptsize{$\pm$ 0.06}}  & \textbf{0.35 \scriptsize{$\pm$ 0.06}} & 0.66 \scriptsize{$\pm$ 0.13} & 0.70 \scriptsize{$\pm$ 0.40}         & 1.45 \scriptsize{$\pm$ 2.24}          \\
		Kin8nm   & 0.09 \scriptsize{$\pm$ 0.00}          & 0.14 \scriptsize{$\pm$ 0.00}             & 0.16 \scriptsize{$\pm$ 0.00}           & 0.13 \scriptsize{$\pm$ 0.01}          & 0.10 \scriptsize{$\pm$ 0.01}  & {\ul 0.08 \scriptsize{$\pm$ 0.00}}    & \textbf{0.07 \scriptsize{$\pm$ 0.00}}    \\
		Naval    & 0.00 \scriptsize{$\pm$ 0.00}             & 0.00 \scriptsize{$\pm$ 0.00}                & 0.00 \scriptsize{$\pm$ 0.00}              & 0.00 \scriptsize{$\pm$ 0.00}                & 0.00 \scriptsize{$\pm$ 0.00}       & 0.00 \scriptsize{$\pm$ 0.00}             & 0.00 \scriptsize{$\pm$ 0.00}                \\
		Power    & 4.11 \scriptsize{$\pm$ 0.17}       & {\ul 3.55 \scriptsize{$\pm$ 0.27}}    & 3.70 \scriptsize{$\pm$ 0.22}         & \textbf{3.35 \scriptsize{$\pm$ 0.15}} & 3.79 \scriptsize{$\pm$ 0.26} & 3.94 \scriptsize{$\pm$ 0.16}       & 3.98 \scriptsize{$\pm$ 0.19}          \\
		Protein  & 4.71 \scriptsize{$\pm$ 0.06}       & \textbf{3.92 \scriptsize{$\pm$ 0.08}} & 4.33 \scriptsize{$\pm$ 0.03}        & {\ul 3.98 \scriptsize{$\pm$ 0.06}}    & 4.79 \scriptsize{$\pm$ 0.52} & 4.32 \scriptsize{$\pm$ 0.03}       & 4.44 \scriptsize{$\pm$ 0.10}           \\
		Wine     & 0.64 \scriptsize{$\pm$ 0.04}       & 0.63 \scriptsize{$\pm$ 0.04}          & 0.62 \scriptsize{$\pm$ 0.04}        & 0.6 \scriptsize{$\pm$ 0.05}           & 0.73 \scriptsize{$\pm$ 0.06} & {\ul 0.44 \scriptsize{$\pm$ 0.09}} & \textbf{0.39 \scriptsize{$\pm$ 0.06}} \\
		Yacht    & 1.58 \scriptsize{$\pm$ 0.48}       & 0.82 \scriptsize{$\pm$ 0.40}           & {\ul 0.50 \scriptsize{$\pm$ 0.20}} & { 0.63 \scriptsize{$\pm$ 0.21}}    & 0.75 \scriptsize{$\pm$ 0.25} & 1.18 \scriptsize{$\pm$ 0.47}       & \textbf{0.47 \scriptsize{$\pm$ 0.11}}           \\
		Year  MSD   & \textbf{8.89 \scriptsize{$\pm$ NA }} & 8.99 \scriptsize{$\pm$ NA }             & {\ul 8.94 \scriptsize{$\pm$ NA }}     & 9.09 \scriptsize{$\pm$ NA }             & 10.97 \scriptsize{$\pm$ NA }   & 9.03 \scriptsize{$\pm$ NA }          & 9.31 \scriptsize{$\pm$ NA }           \\ \midrule
	\end{tabular}
\end{table*}

\paragraph{\textbf{Probabilistic Evaluation.}}

The results in Table \ref{tab:nll} show that TabResFlow consistently performs good, achieving the best or second-best results in seven of nine datasets and the best in four.
Based on average ranks, TabResFlow scores 2.22, outperforming TreeFlow (3.11) and NodeFlow (3.67), with other baselines lagging.
Models like NGBoost and PGBM, which assume specific distributions, struggle on multi-modal datasets such as Protein and Wine.
In contrast, Normalizing Flow models including TreeFlow, NodeFlow, and TabResFlow adapt well to diverse distributions using invertible transformations, resulting in significantly lower likelihood loss.
As shown in Figure \ref{fig:data_distribution}, TabResFlow learns the multi-modal nature of the Protein and Wine datasets.
Ensemble models like CatBoost and Deep.Ens. perform weakest, as they are not optimized for NLL loss.

\paragraph{\textbf{Point Prediction Evaluation.}} 

While TabResFlow is not specifically optimized for point prediction, it achieves competitive RMSE performance, tying with PGBM for most wins in Table \ref{tab:rmse}, demonstrating a strong balance between accuracy and probabilistic modeling. Boosting models like CatBoost, NGBoost, and PGBM outperform NodeFlow and Deep.Ens. on smaller datasets, likely due to their MSE-based training better aligning with RMSE metrics, unlike flow-based models such as TreeFlow, NodeFlow, and TabResFlow. Despite using CatBoost features, TreeFlow lags behind CatBoost in RMSE, possibly due to two-stage training and limited feature encoding. In contrast, TabResFlow’s end-to-end ResNet encoder combined with spline flows enables accurate predictions and robust uncertainty modeling. The high standard deviation of TabResFlow on the energy dataset is caused by three outliers from sampling across 20 cross-validation splits; excluding these yields a result of $0.67 \pm 0.15$.

\begin{table}[t]
	\centering
	\caption{Comparison of TabResFlow with NodeFlow for Inference Time in seconds.}
	\label{tab:time_comparison}
	\begin{tabular}{l|rrc}
		\midrule
		Dataset                    & TabResFlow         & NodeFlow           & Reduction Factor \\ \midrule
		Concrete                   & 5.10 \scriptsize{$\pm$  1.17}     & 15.96 \scriptsize{$\pm$   5.37}   & 3.13           \\
		Energy                     & 2.90 \scriptsize{$\pm$ 0.91}     & 10.94 \scriptsize{$\pm$  1.41}   & 3.77           \\
		Kin8nm                     & 21.16 \scriptsize{$\pm$ 6.73}   & 117.25 \scriptsize{$\pm$ 14.67} & 5.54           \\
		Naval                       & 33.08 \scriptsize{$\pm$ 9.10}    & 162.74 \scriptsize{$\pm$ 33.16} & 4.92           \\
		Power                       & 27.80 \scriptsize{$\pm$ 7.44}    & 154.89 \scriptsize{$\pm$ 41.37} & 5.57           \\
		Protein                     & 132.76 \scriptsize{$\pm$ 20.69} & 727.23 \scriptsize{$\pm$ 118.4} & 5.48           \\
		Wine                        & 5.15 \scriptsize{$\pm$ 1.15}    & 23.60 \scriptsize{$\pm$ 3.51}    & 4.58           \\
		Yacht                      & 1.62 \scriptsize{$\pm$ 0.49}    & 3.96 \scriptsize{$\pm$ 1.17}    & 2.45           \\
		Year MSD                    & 560.87 \scriptsize{$\pm$ NA }    & 8211.45 \scriptsize{$\pm$ NA }   & 14.64           \\ \midrule
	\end{tabular}
\end{table}

\paragraph{\textbf{RunTime Comparisons.}}

TabResFlow achieves the lowest negative log-likelihood while significantly improving runtime efficiency over TreeFlow and NodeFlow, the second- and third-best methods, respectively. Both use CNF components to boost performance, but TreeFlow lacks end-to-end trainability due to its reliance on CatBoost for feature extraction. NodeFlow addresses this with a fully trainable design, making it the primary comparison point in Table \ref{tab:time_comparison}.

On average, TabResFlow is 5.56$\times$ faster than NodeFlow across datasets. NodeFlow and TreeFlow’s CNF modules require solving differential equations during inference, increasing complexity for large datasets like Year MSD and Protein. In contrast, TabResFlow uses neural spline flows with analytical inverses and simple quadratics for faster inference. As Table \ref{tab:ablationTime} shows, even with CNF added, TabResFlow’s ResNet backbone remains more efficient than NodeFlow’s NODE-based architecture.


\subsection{Ablation Experiments}

\begin{table}[t]
	\centering
	\caption{Ablation study on NLL comparing TabResFlow variants: \emph{TabResNet+Gauss} uses a Gaussian head on ResNet; \emph{TabResFlow-Emb} omits numerical MLP embeddings; \emph{TabResFlow+CNF} replaces NSF with CNF flow; and \emph{TabResFlow} includes numerical MLP embeddings with NSF flow.}
	\label{tab:ablationNLL}
	\resizebox{\columnwidth}{!}{%
		\begin{tabular}{l|rrrr}
			\midrule
			Model \textbackslash Dataset            & Concrete         & Energy           & Kin8nm             & Naval \\ \midrule
			NodeFlow           & 3.15 \scriptsize{$\pm$ 0.21}  & 0.90 \scriptsize{$\pm$ 0.25}  & -1.1 \scriptsize{$\pm$ 0.05}    & -5.45 \scriptsize{$\pm$ 0.08}   \\
			TabResNet+Gauss    & 3.07 \scriptsize{$\pm$ 0.31}  & 1.15 \scriptsize{$\pm$ 0.16}  & 0.76 \scriptsize{$\pm$ 0.00}    & 0.74 \scriptsize{$\pm$ 0.00}    \\
			TabResFlow-Emb     & 2.90 \scriptsize{$\pm$ 0.19}   & 0.80 \scriptsize{$\pm$ 0.14}   & -1.17 \scriptsize{$\pm$ 0.05}   & -5.41 \scriptsize{$\pm$ 0.17}   \\
			TabResFlow+CNF & 2.97 \scriptsize{$\pm$ 0.20}  & 1.15 \scriptsize{$\pm$ 0.53}  & -1.29 \scriptsize{$\pm$ 0.04}   & -5.25 \scriptsize{$\pm$ 0.28}   \\
			TabResFlow         & 2.90 \scriptsize{$\pm$ 0.15}   & 0.77 \scriptsize{$\pm$ 0.19}  & -1.29 \scriptsize{$\pm$ 0.04}   & -5.30 \scriptsize{$\pm$ 0.11}    \\ \midrule
		\end{tabular}
	}
\end{table}

Building on the previous section, we examine the key components driving TabResFlow’s performance and efficiency. Due to runtime constraints, experiments focus on the first four datasets (chosen alphabetically) from the nine benchmarks. Tables \ref{tab:ablationNLL} and \ref{tab:ablationTime} compare model variants in terms of NLL and inference time, respectively, with NodeFlow included for reference. Each variant is independently tuned and evaluated over 20 cross-validation splits.

\textit{A1) Are normalizing flow based flexible probabilistic distribution 
	modeling necessary for the given datasets?.}

To explore this, we evaluate a TabResFlow variant \emph{TabResNet+Gauss}, that replaces the RQ-NSF with a Gaussian head for probabilistic prediction. While retaining all other components, it directly learns Gaussian parameters, limiting its ability to capture non-Gaussian distributions. As shown in Table \ref{tab:ablationNLL}, TabResFlow consistently outperforms TabResNet+Gauss across all datasets, with a significant gap on \emph{Naval} (-5.30 vs. 0.74). A similar performance drop is seen when comparing NodeFlow to TabResNet+Gauss. These results demonstrate the critical role of flexible distribution modeling in achieving strong performance, particularly on non-Gaussian data.

\begin{table}[t]
	\centering
	\caption{Ablation study comparing variants of the TabResFlow on inference time in seconds. }
	\label{tab:ablationTime}
	\resizebox{\columnwidth}{!}{%
		\begin{tabular}{l|rrrr}
			\midrule
			Model \textbackslash Dataset            & Concrete         & Energy           & Kin8nm             & Naval \\ \midrule
			NodeFlow           & 15.96 \scriptsize{$\pm$ 5.37} & 10.94 \scriptsize{$\pm$ 1.41} & 117.25 \scriptsize{$\pm$ 14.67} & 162.74 \scriptsize{$\pm$ 33.16} \\
			TabResNet+Gauss    & 0.02 \scriptsize{$\pm$ 0.06}  & 0.02 \scriptsize{$\pm$ 0.06}  & 0.03 \scriptsize{$\pm$ 0.06}    & 0.06 \scriptsize{$\pm$ 0.08}    \\
			TabResFlow-Emb     & 3.1 \scriptsize{$\pm$ 0.95}   & 2.92 \scriptsize{$\pm$ 0.71}  & 20.54 \scriptsize{$\pm$ 6.30}    & 31.28 \scriptsize{$\pm$ 8.09}   \\
			TabResFlow+CNF & 10.75 \scriptsize{$\pm$ 2.30}  & 7.3 \scriptsize{$\pm$ 1.16}  & 79.56 \scriptsize{$\pm$ 9.38}   & 106.98 \scriptsize{$\pm$ 12.01} \\
			TabResFlow         & 5.1 \scriptsize{$\pm$ 1.17}   & 2.9 \scriptsize{$\pm$ 0.91}   & 21.16 \scriptsize{$\pm$ 6.73}   & 33.08 \scriptsize{$\pm$ 9.10}   \\\midrule
		\end{tabular}
	}
\end{table}

\textit{A2) How useful are CNFs compared to RQ-NSF?}. 

To assess the impact of RQ-NSF in TabResFlow, we introduce \emph{TabResFlow+CNF}, which replaces RQ-NSF with a CNF while keeping all other settings unchanged. As shown in Table \ref{tab:ablationNLL}, both variants perform similarly across most datasets, with a notable difference only on \emph{Energy}. However, Table \ref{tab:ablationTime} highlights a key advantage: TabResFlow is over twice as fast during inference. Thus, RQ-NSF matches CNF in performance while offering significantly better efficiency.

\textit{A3) How does the ResNet backbone compare to the NODE architecture as a feature extractor?.} 

NodeFlow uses a NODE-based backbone, while TabResFlow relies on a simpler ResNet. Since both NodeFlow and TabResFlow+CNF share a CNF for probabilistic modeling, their comparison in Tables~\ref{tab:ablationNLL} and \ref{tab:ablationTime} isolates the impact of the backbone. TabResFlow+CNF performs better on \emph{Concrete} and \emph{Kin8nm}, while NodeFlow leads on the other two datasets. However, ResNet proves far more efficient, achieving faster inference across the board. Combined with RQ-NSF, the ResNet backbone allows TabResFlow to outperform NodeFlow overall. In summary, ResNet offers similar predictive accuracy with significantly better efficiency.

\textit{A4) How useful are the numerical embeddings?.}

Finally, we assess the effect of numerical embeddings by comparing TabResFlow with TabResFlow-Emb, which omits the MLP-based embedding of numerical features. As shown in Table \ref{tab:ablationNLL}, embedding generally improves performance across datasets, with the exception of \emph{Naval}, where suboptimal hyperparamter tuning may explain the difference. Prior works \cite{gorishniy2022embeddings,somepalli2021saint} also supports the benefit of MLP or piecewise embeddings for numerical inputs. Together, the use of numerical embeddings, a ResNet backbone, and spline-based flows contributes to TabResFlow’s strong accuracy and efficiency in likelihood estimation.

The next section examines the role of probabilistic prediction in the selective prediction framework \cite{geifman2017selective}, introducing the Area Under Risk-Coverage (AURC) as an effective metric to evaluate uncertainty in regression tasks


\section{Case Study: TabResFlow for Probabilistic Used-Car pricing}

Beyond evaluations on public datasets, we apply the proposed TabResFlow to a \emph{large-scale} \emph{real-world industry} use case: \emph{used-car pricing}. With the growing importance of online marketplaces and rising demand for used cars in a fast evolving automotive landscape, rapid, adaptive, and precise pricing is essential for both consumers and dealers.
Automating fair pricing for used cars necessitates a machine learning model that is capable of quantifying the uncertainty associated with price estimates, or in other words, predicting the distribution of used-car prices. As outlined in section \ref{sec:problem}, this task translates into a probabilistic univariate regression problem with a single output feature, representing the expected price distribution based on vehicle attributes.

\begin{figure*}[htbp]
	\begin{tabular}{ccc}  
		\label{fig:Dec2023}{\includegraphics[width=0.31\linewidth]{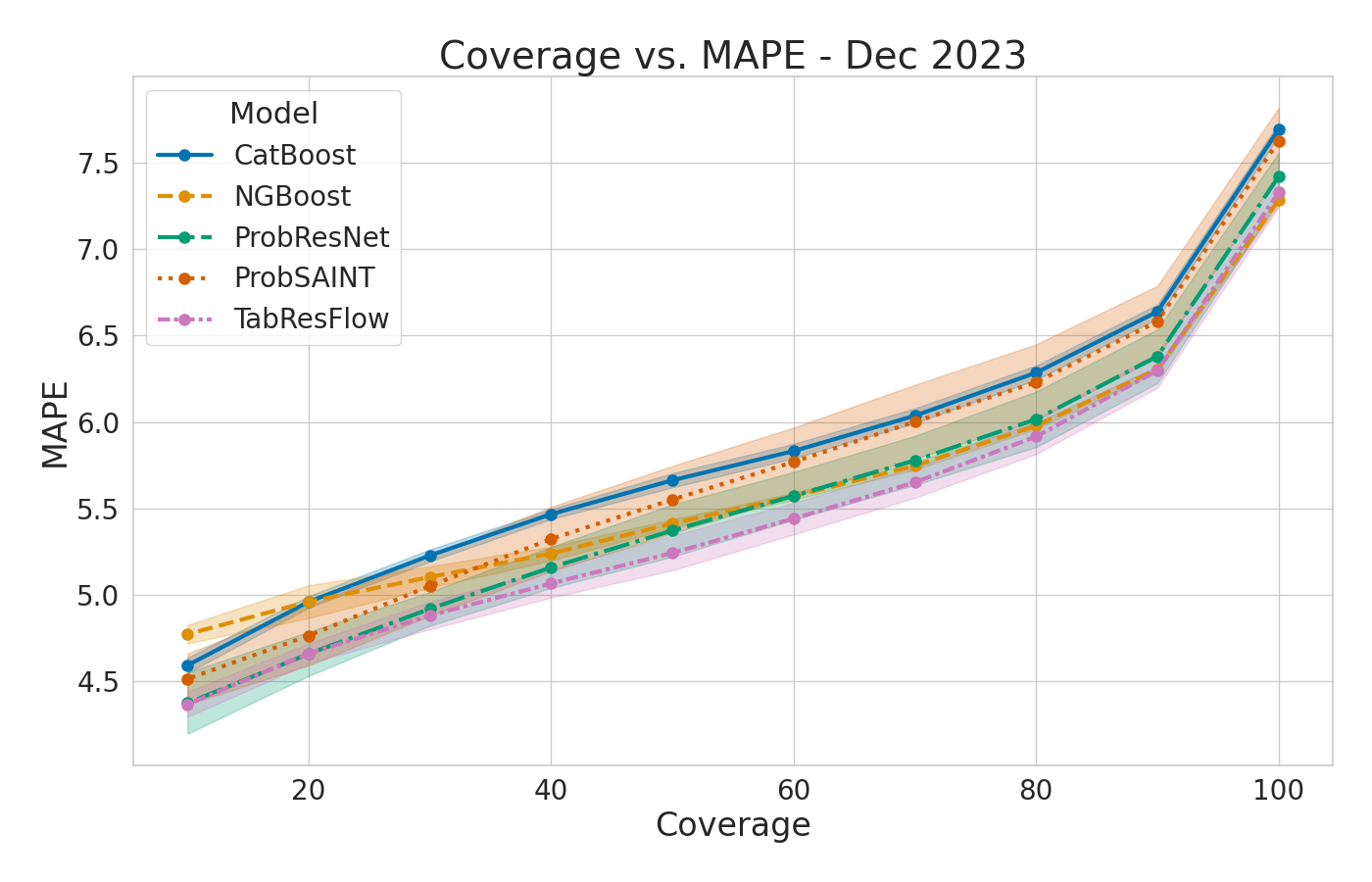}} &
		\label{fig:Jan2024}{\includegraphics[width=0.31\linewidth]{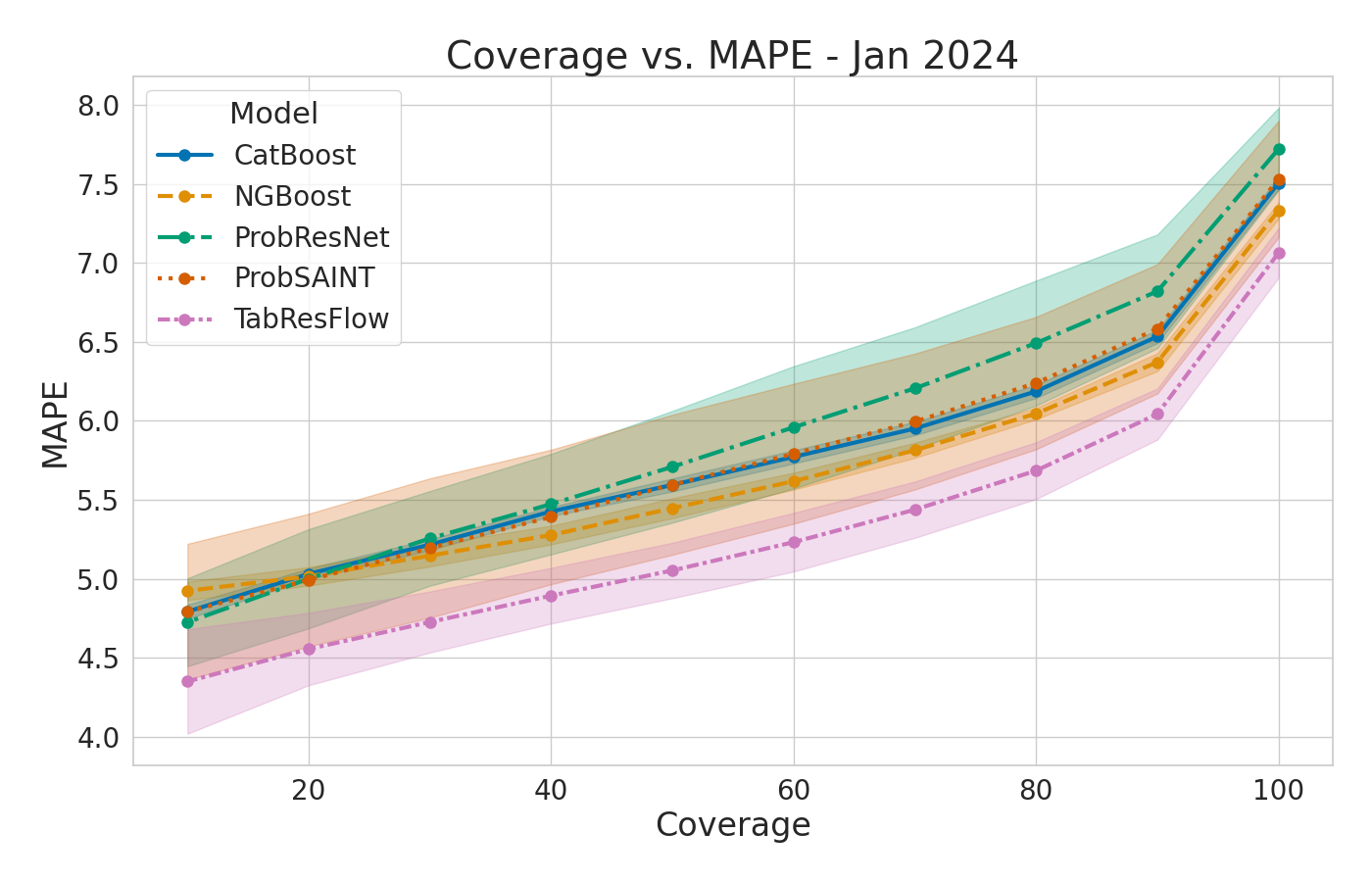}} & \label{fig:Feb2024}{\includegraphics[width=0.31\linewidth]{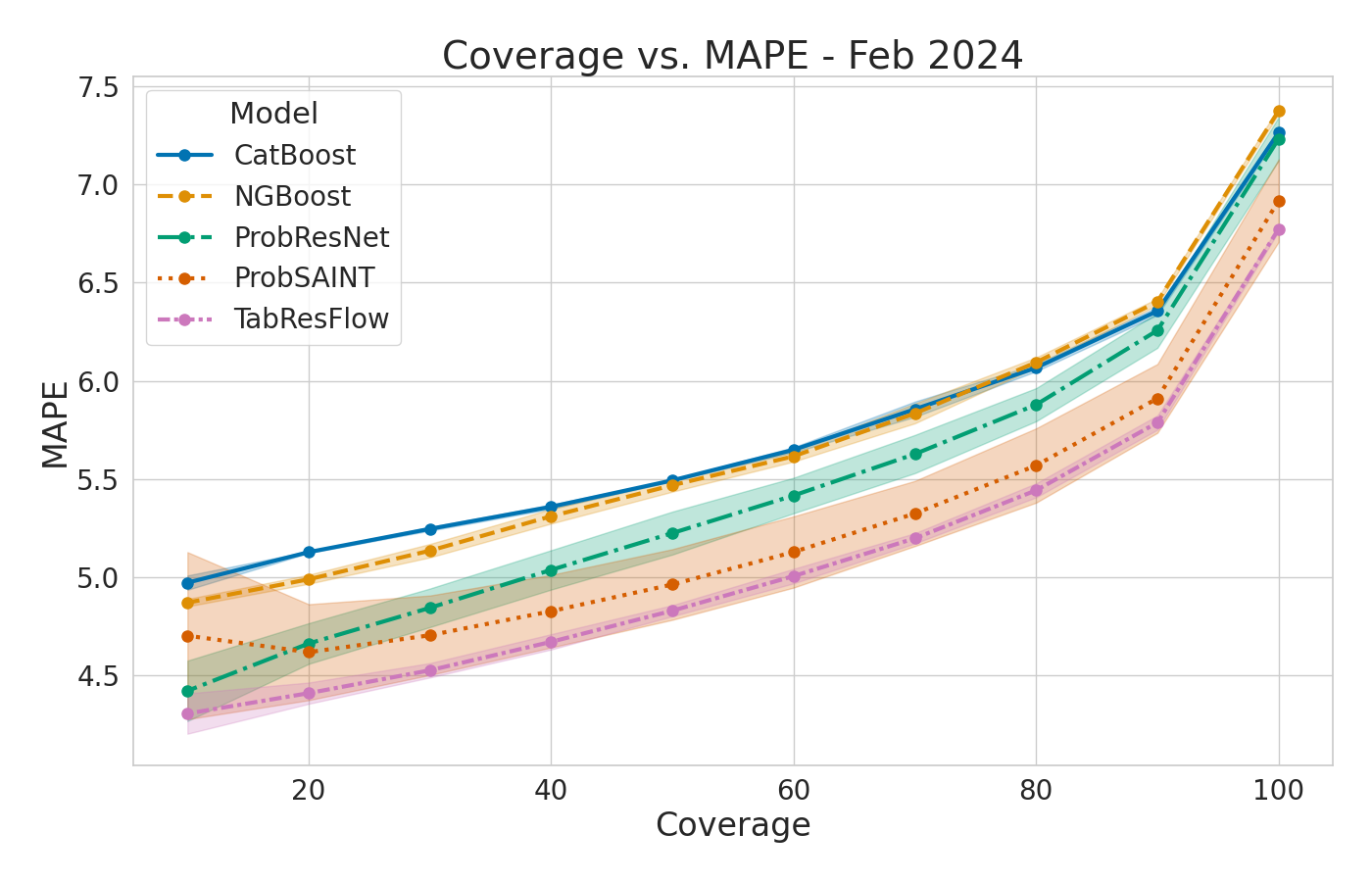}}\\
	\end{tabular}%
	\caption{Risk-Coverage Plot. Comparison of different probabilistic models with proposed TabResFlow. 
		x-axis denotes the coverage ordered by increasing order of confidence and y-axis denotes the risk in terms of 
		MAPE error at each coverage.}
	\label{fig:riskcoverage}
	\vspace{1em}
\end{figure*}

\paragraph{\textbf{Dataset.}} The used-car pricing dataset, sourced externally and standardized from real-world Business-to-Customer (B2C) online platforms such as autoscout24 and mobile.de, comprises 64 features detailing vehicle attributes, including type, brand, and registration year alongside usage metrics such as damage history, offer duration, and mileage. These features span categorical, numerical, and date-based data types, with key variables and their corresponding datatypes listed in Table~\ref{tbl:dataset_feat} of the supplementary material.
The dataset, spanning February 2022 to February 2024, contains 2,239,473 records, 4.34 times larger than the biggest public YearMSD dataset in the previous section. Missing values are handled by assigning an "empty" category for categorical variables and "-1" for numerical ones. This large-scale experiments offers key insights into TabResFlow’s scalability for real-world applications.

\paragraph{\textbf{Experimental Setup.}} We adopt a chronological splitting strategy similar to \cite{madhusudhanan2024probsaint}, dividing the dataset into training, validation, and test sets based on the sales date. Specifically, the last four weeks of sales data serve as the test set, while the four weeks preceding the test period function as the validation set, with all prior data used for training.
For instance, the \emph{Feb-2024} dataset includes the final four weeks leading up to February 2024 as test data, January 2024 as validation data, and all earlier records for training. This approach mirrors real-world conditions, where pricing predictions rely on historical transactions up to the preceding week.
To ensure experimental robustness, we generate three dataset variants with different testing periods: \emph{Dec-2023}, \emph{Jan-2024}, and \emph{Feb-2024}. The precise split ranges and sample counts are detailed in Table~\ref{tbl:company_dataset_split} of the supplementary material.

\subsection{Evaluation: Area under Risk-Coverage}
In mission-critical settings, reliable probabilistic predictions prioritize high-confidence outputs while tolerating uncertainty in others. Greater coverage of test examples at high confidence is crucial; for example, lower error at 50\% coverage is often preferred over lower error at 40\% coverage. Thus, balancing Risk/Error and Coverage is a key metric alongside measures like NLL.

Selective Prediction research closely aligns with this area, utilizing Risk-Coverage graphs as an established method for evaluating probabilistic predictions \cite{geifman2017selective}. Recognizing the significance of these plots in assessing model uncertainty, here, we suggest a method to  
(1) derive uncertainty scores from models, (2) convert them to confidence scores, and (3) generate Risk-Coverage graphs, defining the Area under Risk-Coverage (AURC) as the area beneath these curves.

\paragraph{\textbf{Confidence Score Calculation.}} In a Gaussian distribution, the deviation from the mean value is commonly used to quantify uncertainty, typically represented by the standard deviation ($\sigma$) or variance ($\sigma^2$). For instance, the negative variation from the predicted mean has been proposed as an effective confidence metric \cite{geifman2017selective}. For our experiments, we define confidence score as the inverse of the standard deviation ($\sigma$) i.e, 	
$$\text{Conf.} = \frac{1}{\sigma}$$

\paragraph{\textbf{Risk-Coverage Plots.}} Risk-Coverage plots illustrate the trade-off between Risk/Error and Coverage, providing a metric for evaluating probabilistic model performance. In these plots, data points are ranked in descending order based on their confidence scores, ensuring that the most confident 10\% of examples constitute the 10\% coverage. The corresponding error for this subset is then plotted on the Y-axis, as depicted in Figure \ref{fig:riskcoverage}. 

\paragraph{\textbf{Area under Risk-Coverage}}
We compute the Area Under Risk-Coverage (AURC) from the Risk-Coverage plot at 10\% coverage intervals, applying the trapezoidal rule for approximation. Given a set of coordinates ($u_c,v_c$) from the Risk-Coverage plot, where $u_c$ represents the coverage and $v_c$ represents the average error, for different coverage levels $c$, the area under the curve can be approximated as

\begin{align*}
	\text{AURC} \approx \frac{1}{100} \sum_{c \in \{20, 30, \dots, 100\}} \frac{v_c + v_{c-10}}{2} (u_{c} - u_{c-10})
\end{align*}

In the context of used-car pricing, the Mean Absolute Percentage Error (MAPE) is the preferred metric over unnormalized errors such as MAE or RMSE. Given its relevance in business applications, we employ MAPE as the Risk/Error measure for AURC calculation.

\paragraph{\textbf{Baselines.}} We compare TabResFlow with relevant baselines in Table \ref{tab:AURC}, including CatBoost with ensemble-based uncertainty \cite{malinin2020uncertainty}, NGBoost using Natural Gradients for probabilistic predictions \cite{duan2020ngboost}, and ProbSAINT, which learns a Gaussian distribution atop a SAINT architecture \cite{madhusudhanan2024probsaint,somepalli2021saint}. ProbSAINT demonstrated superior NLL and MAPE performance on five years of used-car pricing data \cite{madhusudhanan2024probsaint}. We also compare against a variant ProbResNet, with TabResFlow’s ResNet backbone but with a Gaussian head instead of Neural Spline Flows \cite{madhusudhanan2024probsaint}. Due to inconsistencies between CPU and GPU versions, we exclude PGBM results on the used-car dataset.

\paragraph{\textbf{Results.}} 

The results in Table \ref{tab:AURC} and the Risk-Coverage plot in Figure \ref{fig:riskcoverage} show that TabResFlow achieves the highest score in the probabilistic AURC metric. Overall, deep learning models outperform boosting methods in large-scale probabilistic tabular tasks. ProbSAINT, ProbResNet, and TabResFlow also perform on par with CatBoost and NGBoost on MAPE score. Among boosting methods, NGBoost consistently surpasses CatBoost in both AURC and MAPE. In the Feb-2024 dataset, NGBoost records an AURC of 5.10 vs. CatBoost’s 5.13, while CatBoost achieves a lower MAPE (7.26 vs. 7.37). This indicates NGBoost models uncertainty more effectively, despite CatBoost’s stronger point predictions. A key drawback of NGBoost, however, is its lack of GPU support, which hampers scalability for large datasets. 
Comparing deep learning backbones, SAINT generally outperforms ResNet in MAPE except in Dec-2023, where ResNet leads. However, SAINT’s dual attention mechanism incurs substantial computational cost, making integration with normalizing flows less practical for large datasets due to increased complexity.

Assessing the impact of normalizing flows, TabResFlow significantly improves upon ProbResNet, sharing the same ResNet backbone but outperforming notably in AURC. This gain is attributed to the flexible distribution modeling enabled by Neural Spline Flows versus Gaussian assumptions in baseline models. On the Dec-2023 dataset, TabResFlow achieves the top AURC score, outperforming NGBoost in uncertainty quantification despite NGBoost’s better MAPE, underscoring TabResFlow’s strength in probabilistic modeling on large-scale data.

\begin{table}[t]
	\caption{Comparison on TabResFlow with other baselines on three variants of the used-car dataset. Probabilistic metric `AURC'
		indicates the Area Under Risk-Coverage curve and point-prediction result `MAPE' is the Mean Absolute Percentage Error. 
		The best method is in \textbf{bold} and the second best method is \underline{underlined}.}
	\label{tab:AURC}
	\centering
	\resizebox{\columnwidth}{!}{%
		\begin{tabular}{l|ccc|ccc}
			\midrule
			& \multicolumn{3}{c|}{AURC}                                                                   & \multicolumn{3}{c}{MAPE}                                                                   \\ 
			& \multicolumn{1}{c}{Feb-2024} & \multicolumn{1}{c}{Jan-2024} & \multicolumn{1}{c|}{Dec-2023} & \multicolumn{1}{c}{Feb-2024} & \multicolumn{1}{c}{Jan-2024} & \multicolumn{1}{c}{Dec-2023} \\ \midrule
			CatBoost   & 5.13$\pm$0.01              & 5.19$\pm$0.03              & 5.23$\pm$0.01              & 7.26$\pm$0.01              & 7.51$\pm$0.04              & 7.68$\pm$0.03              \\
			NGBoost    & 5.10$\pm$0.02               & \underline{5.09$\pm$0.04}              & 5.03$\pm$0.02              & 7.37$\pm$0.01              & \underline{7.38$\pm$0.13}              & \textbf{7.28$\pm$0.03}              \\
			ProbSAINT  & \underline{4.69$\pm$0.15}              & 5.19$\pm$0.34              & 5.13$\pm$0.15              & \underline{6.92$\pm$0.21}              & 7.39$\pm$0.15              & 7.63$\pm$0.19              \\
			ProbResNet & 4.88$\pm$0.07              & 5.31$\pm$0.27              & \underline{4.98$\pm$0.11}              & 7.26$\pm$0.03              & 7.66$\pm$0.28              & 7.40$\pm$0.13               \\
			TabResFlow    & \textbf{4.54$\pm$0.03}              & \textbf{4.73$\pm$0.15}              & \textbf{4.90$\pm$0.06}               & \textbf{6.85$\pm$0.11}              & \textbf{7.06$\pm$0.16}              & \underline{7.32$\pm$0.08}            \\\midrule
		\end{tabular}%
	}
\end{table}
\section{Conclusion}

We present TabResFlow, a computationally efficient method for probabilistic univariate tabular regression capable of modeling non-Gaussian, multimodal distributions. By incorporating monotonic rational quadratic splines, TabResFlow enables exact log-likelihood computation with a tractable Jacobian.
Experiments on nine benchmark datasets validate its effectiveness in probabilistic NLL and RMSE metrics, further supported by ablation studies assessing key components. To demonstrate real-world scalability, we apply TabResFlow to selective used-car pricing, introducing Area Under Risk Coverage (AURC) as a practical metric for probabilistic model evaluation. Results confirm TabResFlow’s robustness, outperforming baselines in AURC and MAPE metrics.

\paragraph{\textbf{Limitations.}} A major limitation of our analysis of TabResFlow is the absence of evaluation in the multivariate setting. Although applying TabResFlow to multivariate cases is straightforward, our study focused on univariate regression, where standard normalizing flows such as RealNVP and MAF methods are ineffective. Additionally, while we introduced the inverse standard deviation as an uncertainty metric, exploring entropy-based uncertainty measures remains necessary. Finally, to better understand the scenarios where TabResFlow is most beneficial, conducting scaling experiments with varying dataset sizes would be valuable.



\bibliography{mybibfile}

\clearpage
\section{Supplementary Material}

\subsection{Public Dataset Experiments}

\subsubsection{Dataset Characteristics}
\label{app:summary_stats_public}
\begin{table}[h]
	\centering
	\caption{Dataset Characteristics. N - denotes the number of datapoints, CV Splits - denotes the number of 
		cross-validation splits, D - Input Features and P - target dimension (Univariate)}
	\begin{tabular}{lrcrr}
		\toprule
		\textbf{Dataset} & \textbf{N} & \textbf{CV Splits} & \textbf{D} & \textbf{P} \\
		\midrule
		Concrete & 1030 & 20 CV & 8 & 1 \\
		Energy & 768 & 20 CV & 8 & 1 \\
		Kin8nm & 8192 & 20 CV & 8 & 1 \\
		Naval & 11,934 & 20 CV & 16 & 1 \\
		Power & 9568 & 20 CV & 4 & 1 \\
		Protein & 45,730 & 5 CV & 9 & 1 \\
		Wine & 1588 & 20 CV & 11 & 1 \\
		Yacht & 308 & 20 CV & 6 & 1 \\
		Year MSD & 515,345 & 1 CV & 90 & 1 \\
		\bottomrule
	\end{tabular}
	\label{tab:dataset_characteristics}
\end{table}

\subsubsection{CRPS scores - public dataset}
\label{app:crps_public}
\begin{table}[h]
	\caption{Comparison of probabilisitic tabular regression models on benchmark datasets for CRPS score. The best value is represented in \textbf{bold}.}
	\label{tab:crps}
	\centering
	\begin{tabular}{l|rrr}
		\midrule
		& CatBoost                 & NodeFlow                 & TabResFlow               \\ \midrule
		Concrete & \textbf{2.24 $\pm$ 0.36} & 2.80 $\pm$ 0.34           & 2.52 $\pm$ 0.3           \\
		Energy   & \textbf{0.23 $\pm$ 0.03} & 0.35 $\pm$ 0.14          & 0.43 $\pm$ 0.29          \\
		Kin8nm   & 0.06 $\pm$ 0.00             & \textbf{0.04 $\pm$ 0.00}    & \textbf{0.04 $\pm$ 0.00}    \\
		Power    & \textbf{1.68 $\pm$ 0.11} & 1.95 $\pm$ 0.06          & 1.96 $\pm$ 0.07          \\
		Protein  & 1.90 $\pm$ 0.11           & \textbf{1.75 $\pm$ 0.03} & 1.78 $\pm$ 0.04          \\
		Wine     & \textbf{0.34 $\pm$ 0.02} & \textbf{0.34 $\pm$ 0.02} & \textbf{0.34 $\pm$ 0.04} \\
		Yacht    & \textbf{0.41 $\pm$ 0.24} & 0.50 $\pm$ 0.19           & 0.56 $\pm$ 0.21          \\
		Year     & -                        & 4.22 $\pm$ 0.00             & 4.40 $\pm$ 0.00           \\ \midrule  
	\end{tabular}
\end{table}

\subsubsection{Area under Risk Coverage - Public Datasets}

\begin{table}[t]
	\centering
	\caption{Comparison of TabResFlow with NodeFlow for AURC metric. 'AURC-Entropy' indicates the confidence score based on Entropy as the uncertainty measure and RMSE as the evaluation measure. Similarly, 'AURC-Std. Dev.' is the confidence score based on deviation from the mean of the predicted distribution and RMSE as the evaluation metric.}
	\label{tab:aurcpublic}
	\begin{tabular}{l|cc|cc}
		\midrule
		& \multicolumn{2}{c|}{AURC-Entropy} & \multicolumn{2}{c}{AURC-Std. Dev.} \\ \midrule
		& \multicolumn{1}{l}{NodeFlow} & \multicolumn{1}{l|}{TabResFlow} & \multicolumn{1}{l}{NodeFlow} & \multicolumn{1}{l}{TabResFlow} \\ \midrule
		Concrete & 4.27            & 4.04           & 3.86             & 3.63            \\
		Energy   & 0.53            & 1.67           & 0.42             & 1.61            \\
		Kin8nm   & 0.07            & 0.06           & 0.06             & 0.05            \\
		Power    & 3.29            & 3.49           & 2.97             & 2.97            \\
		Protein  & 0.32            & 0.31           & 0.27             & 0.26            \\
		Wine     & 0.57            & 0.62           & 0.54             & 0.57            \\
		Yacht    & 0.88            & 0.41           & 0.58             & 0.37            \\
		Year     & 0.32            & 0.36           & 0.26             & 0.26           \\ \midrule
	\end{tabular}
\end{table}

Table \ref{tab:aurcpublic} compares AURC score for confidence scores based on Entropy - InvEntropyRMSE and 
confidence scores based on Standard deviation - InvStdRMSE for error metric RMSE as the y-axis. We 
report the results on four random datasets from the nine benchmark datasets.

\begin{figure}[t]
	\centering
	\begin{tabular}{cc}
		\begin{minipage}{0.45\linewidth}
			\centering
			\includegraphics[width=\linewidth]{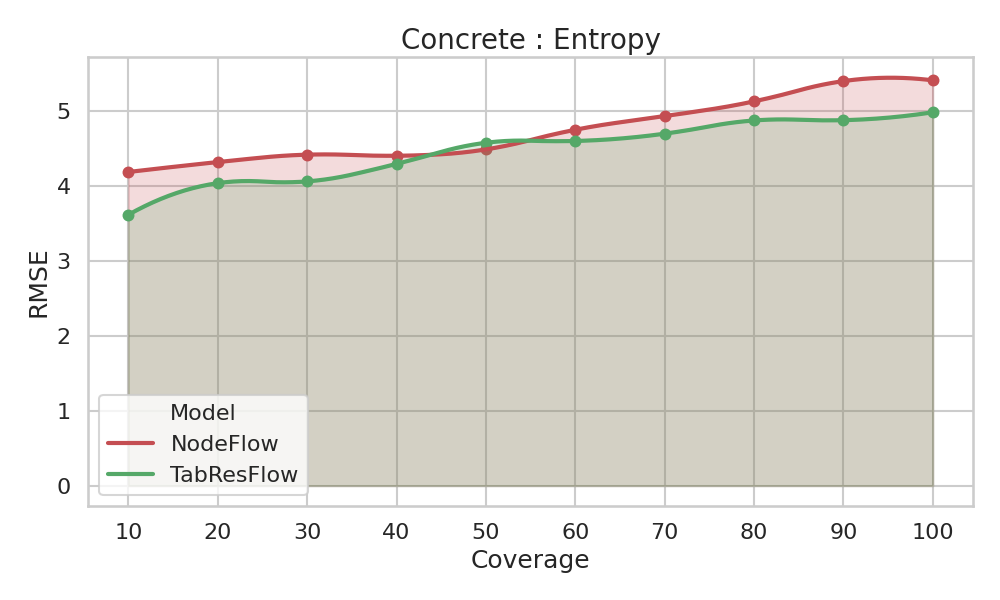}
			\\ (a)
		\end{minipage}
		&
		\begin{minipage}{0.45\linewidth}
			\centering
			\includegraphics[width=\linewidth]{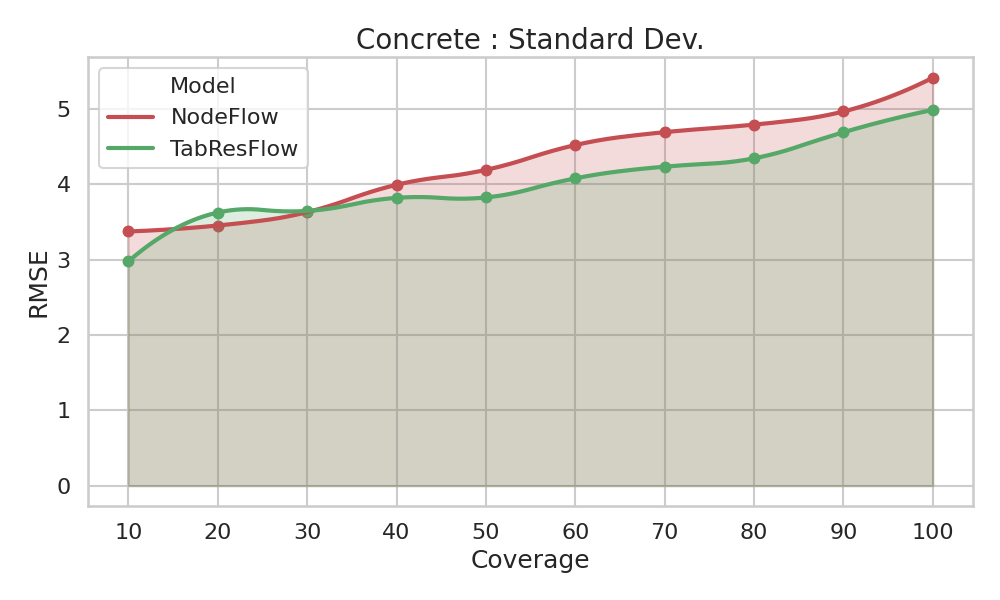}
			\\ (b)
		\end{minipage}
		\\
		\\[-1ex]
		\begin{minipage}{0.45\linewidth}
			\centering
			\includegraphics[width=\linewidth]{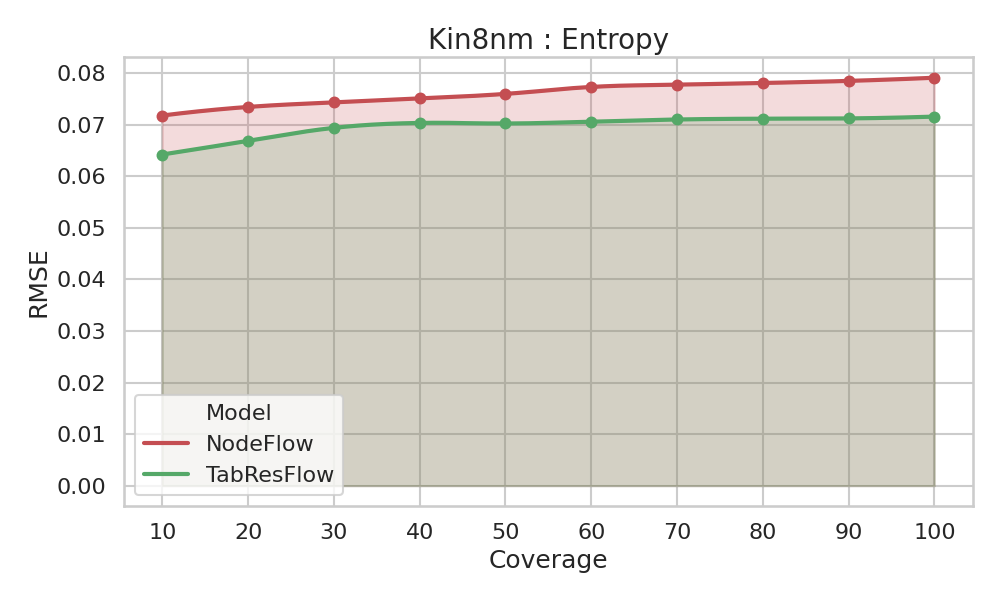}
			\\ (c)
		\end{minipage}
		&
		\begin{minipage}{0.45\linewidth}
			\centering
			\includegraphics[width=\linewidth]{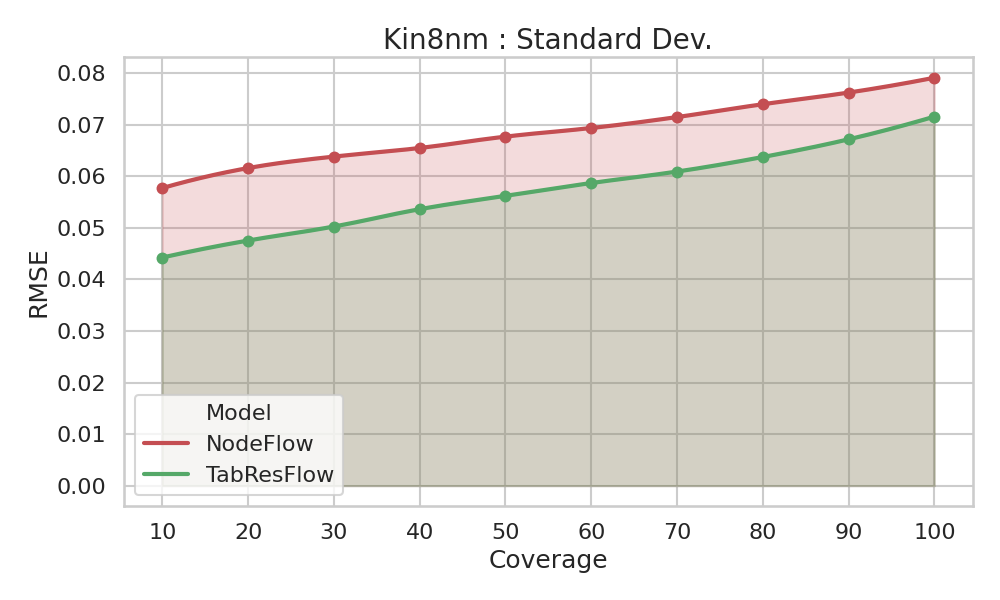}
			\\ (d)
		\end{minipage}
	\end{tabular}
	\vspace{2em}
	\caption{Risk–Coverage curves on public datasets. (a) and (b): Concrete dataset using Entropy and Standard deviation as uncertainty measures, respectively. (c) and (d): Corresponding plots for the Kin8nm dataset.}
	\label{fig:publicRiskCoverage}
\end{figure}

\paragraph{\text{Results.}} Figure \ref{fig:publicRiskCoverage} compares the Risk-Coverage plot for 
TabResFlow and NodeFlow models on \emph{Concrete} and \emph{Kin8nm} dataset utilizing the entropy and the standard deviation 
of the predicted distribution as the uncertainty metric respectively. The inverse of the 
chosen uncertainty measure is used to quantify the model confidence on a particular sample. 
Firstly, TabResFlow and NodeFlow, both provide approximately monotonic plots such that as 
coverage increases so does the Risk/RMSE error. However, as seen from the Figures \ref{fig:publicRiskCoverage}
and the Table \ref{tab:aurcpublic}, TabResFlow provides better Risk-Coverage in comparison to NodeFlow on 
the Concrete dataset. 

From the Table \ref{tab:aurcpublic}, we observe that both inverted-entropy and inverted-standard deviation 
are suitable confidence scores for a normalizing flow model. It is interesting to observe that 
the inverted-standard deviation provides lower AURC in comparison to inverted-entropy across the five datasets.
This suggests that inverted-standard deviation is a useful confidence metric even for non-gaussian distributions.
Another interesting observation is that having a lower negative loglikelihood loss does not guarantee a better 
AURC score. For example, on NLL scores, for both Power and Wine datasets, the TabResFlow outperforms the NodeFlow 
model. However, this is not the case with the AURC scores where, NodeFlow is able to outperform TabResFlow. 


\subsection{Used Car Pricing}

\subsubsection{Dataset Description}
\label{app:data_usedcar_description}

Table \ref{tbl:dataset_feat} shows the dataset features and corresponding datatypes. Whereas Table \ref{tbl:company_dataset_split} illustrates the dataset splits that were employed.
\begin{table}[h]
	\centering
	\caption{Example used-car features and corresponding data types. The ``sales price'' is our target feature.}
	\label{tbl:dataset_feat}
	\begin{tabular}{ll|ll}
		\hline
		\textbf{Feature}     & \textbf{Type} & \textbf{Feature}        & \textbf{Type} \\
		\hline
		Odometer             & Numeric       & Available Date         & Date        \\
		Condition            & Categorical        & Engine Type             & Categorical        \\
		Age                  & Integer       & Engine Size             & Numeric       \\
		Offer Duration           & Integer       & Engine Power            & Numeric       \\
		Make                 & Categorical        & Fuel Type               & Categorical        \\
		Initial Registration & Date          & Transmission            & Categorical        \\
		Model                & Categorical        & Euro Emissions Std & Categorical       \\ 
		Model Variant        & Categorical        & Sales Date & Date       \\ 
		Brand                & Categorical        & Sales Price & Numeric       \\ 
		\hline
	\end{tabular}
\end{table}

\begin{table*}[t]
	\caption{Comparison on TabResFlow with other baselines on three variants of the used-car dataset. Probabilistic metric `NLL' and `CRPS'
		and point-prediction result `RMSE'. 
		The best method is in \textbf{bold}.}
	\label{tab:usedcarNLL}
	\centering
	\resizebox{\linewidth}{!}{%
		\begin{tabular}{l|rrr|rrr|rrr}
			\midrule
			\textbf{Metric}                        & \multicolumn{3}{c|}{\textbf{NLL}}                                                           & \multicolumn{3}{c|}{\textbf{CRPS}}                                            & \multicolumn{3}{c}{\textbf{RMSE}}                                             \\ \midrule
			\textbf{Models\textbackslash{}Dataset} & \multicolumn{1}{c}{Feb-2024} & \multicolumn{1}{c}{Jan-2024} & \multicolumn{1}{c|}{Dec-2023} & Feb-2024                & Jan-2024                 & \multicolumn{1}{c|}{Dec-2023}                & Feb-2024                 & Jan-2024                 & \multicolumn{1}{c}{Dec-2023}                \\ \midrule
			\textbf{CatBoost}                      & 8.29 (0.00)                  & 8.31 (0.00)                  & 8.80 (0.01)                  & \textbf{1402.19 (3.54)} & 1441.50 (7.64)           & 1599.00 (6.35)          & {3229.31 (4.98)}  & {3344.18 (13.83)} & 4378.73 (10.71)         \\
			\textbf{NGBoost}                       & 8.31 (0.01)                  & 8.32 (0.01)                  & \textbf{8.30 (0.01)}          & 1402.52 (3.44)          & \textbf{1409.73 (24.37)} & \textbf{1390.19 (4.77)} & 3239.12 (3.83)           & \textbf{3260.82 (55.83)} & \textbf{3232.65 (8.76)} \\
			\textbf{ProbResNet}                    & \textbf{8.25 (0.01)}         & 8.34 (0.06)                  & 8.72 (0.03)                  & 1429.00 (8.18)          & 1535.06 (63.25)          & 1564.6 (29.31)          & 3367.23 (87.39)          & 3571.67 (128.57)         & 4363.76 (59.74)         \\
			\textbf{ProbSAINT}                     & 8.38 (0.29)                  & \textbf{8.29 (0.02)}         & 8.68 (0.04)                  & 1333.63 (34.82)         & 1475.42 (37.53)          & 1611.86 (39.73)         & \textbf{3152.23 (63.01)} & 3425.57 (69.59)          & 4455.95 (62.16)         \\
			\textbf{TabResFlow}                       & 9.03 (0.01)                  & 9.08 (0.02)                  & 9.16 (0.01)                  & 1395.4 (26.55)          & 1432.93 (26.74)          & 1630.79 (21.00)            & 3489.18 (42.00)             & 3588.7 (28.15)           & 4659.92 (46.18)    \\\midrule    
		\end{tabular}
	}
\end{table*}

\subsubsection{Dataset Split}
\label{app:data_usedcar_split}
\begin{table}[]
	\centering
	\caption{Used-Car dataset split for train validation and test for the three dataset variants. Dates in 
		'YYYY-MM-DD' format.}
	\label{tbl:company_dataset_split}
	\begin{tabular}{r|l|ccc}
		\midrule
		\multicolumn{1}{l|}{Dataset} & Feat.   & Train End  & Validation End & Test End   \\ \midrule
		\multirow{2}{*}{Feb-2024}   & Date    & 2023-12-26 & 2024-01-23     & 2024-02-20 \\
		& Samples & 2,059,324    & 81,525          & 98,624      \\ \midrule
		\multirow{2}{*}{Jan-2024}   & Date    & 2023-11-28 & 2023-12-26     & 2024-01-23 \\
		& Samples & 1,980,417    & 78,907          & 81,525      \\ \midrule
		\multirow{2}{*}{Dec-2023}   & Date    & 2023-10-31 & 2023-11-28     & 2023-12-26 \\
		& Samples & 1,888,215    & 92,202          & 78,907     \\  \midrule
	\end{tabular}
\end{table}

\subsubsection{Additional Metrics : UsedCar Pricing}

We evaluate other additional metrics for the used car pricing dataset and report these results in the Table \ref{tab:usedcarNLL}.

\end{document}